\def\year{2020}\relax
\documentclass[letterpaper]{article} % DO NOT CHANGE THIS
\usepackage{aaai20}  % DO NOT CHANGE THIS
\usepackage{times}  % DO NOT CHANGE THIS
\usepackage{helvet} % DO NOT CHANGE THIS
\usepackage{courier}  % DO NOT CHANGE THIS
\usepackage[hyphens]{url}  % DO NOT CHANGE THIS
\usepackage{graphicx} % DO NOT CHANGE THIS
\usepackage{graphicx}  % DO NOT CHANGE THIS

\usepackage{epsfig}
\usepackage{amsmath}
\usepackage{amssymb}
\usepackage{soul}
\usepackage[utf8]{inputenc}
\usepackage[small]{caption}
\usepackage{booktabs}
\usepackage[ruled,lined,linesnumbered]{algorithm2e}
\usepackage{algorithmic}
\usepackage{xcolor}
\usepackage{adjustbox}
\usepackage{subfigure}
\usepackage{blindtext}
\usepackage{enumerate}
\usepackage{bm}
\usepackage{dsfont}
\usepackage{multirow}
\usepackage[switch]{lineno}

\newenvironment{packed_itemize}{
\begin{itemize}
  \setlength{\itemsep}{1pt}
  \setlength{\parskip}{0pt}
  \setlength{\parsep}{0pt}
}{\end{itemize}}

\title{Multi-source Distilling Domain Adaptation}
\author{
Sicheng Zhao$^{1}$\thanks{Corresponding Author. $^{\#}$ Equal Contribution.}$^{\#}$, Guangzhi Wang$^{2\#}$, Shanghang Zhang$^{1\#}$, Yang Gu$^{2}$, Yaxian Li$^{23}$, \\
\Large \textbf{Zhichao Song$^{2}$, Pengfei Xu$^{2}$, Runbo Hu$^{2}$, Hua Chai$^{2}$, Kurt Keutzer$^{1}$}\\
$^{1}$University of California, Berkeley, USA  $^{2}$Didi Chuxing, China  $^{3}$Renmin University of China, China\\
  schzhao@gmail.com, gzwang98@gmail.com, shzhang.pku@gmail.com, liyaxian@ruc.edu.cn\\
  \{guyangdavid,songzhichao,xupengfeipf,hurunbo,chaihua\}@didiglobal.com, keutzer@berkeley.edu \\
}
 
\begin{document}

\maketitle

\begin{abstract}
Deep neural networks suffer from performance decay when there is domain shift between the labeled source domain and unlabeled target domain, which motivates the research on domain adaptation (DA). Conventional DA methods usually assume that the labeled data is sampled from a single source distribution. However, in practice, labeled data may be collected from multiple sources, while naive application of the single-source DA algorithms may lead to suboptimal solutions.
In this paper, we propose a novel multi-source distilling domain adaptation (MDDA) network, which not only considers the different distances among multiple sources and the target, but also investigates the different similarities of the source samples to the target ones. Specifically, the proposed MDDA includes four stages: (1) pre-train the source classifiers separately using the training data from each source; (2) adversarially map the target into the feature space of each source respectively by minimizing the empirical Wasserstein distance between source and target; (3) select the source training samples that are closer to the target to fine-tune the source classifiers; and (4) classify each encoded target feature by corresponding source classifier, and aggregate different predictions using respective domain weight, which corresponds to the discrepancy between each source and target. Extensive experiments are conducted on public DA benchmarks, and the results demonstrate that the proposed MDDA significantly outperforms the state-of-the-art approaches. Our source code is released at: \url{https://github.com/daoyuan98/MDDA}.
\end{abstract}

\begin{figure}[t]
   \centering
   \includegraphics[width=0.95\linewidth]{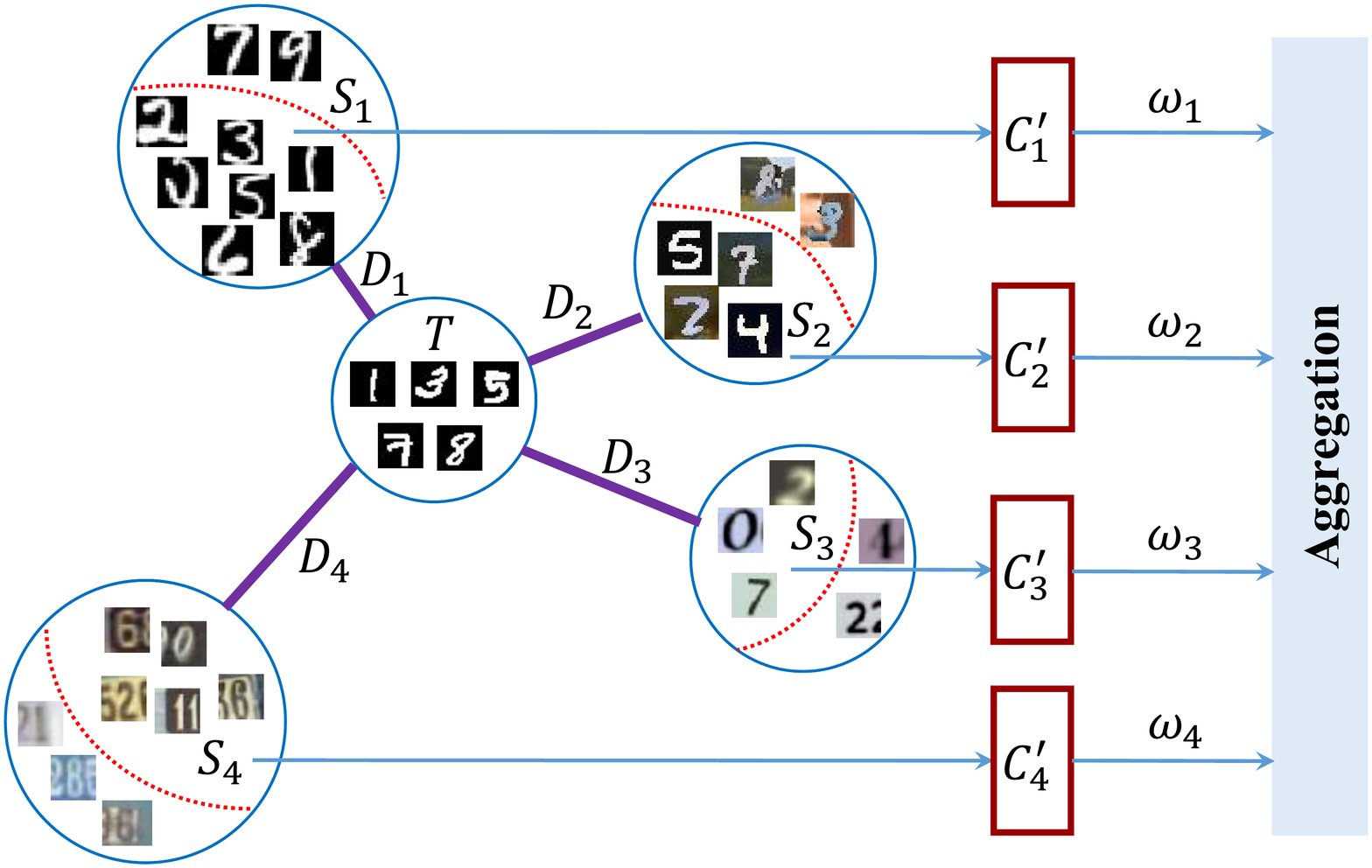}
   \caption{Illustration of MDDA which explores the relationships among different sources and the target. We employ a discriminator $D$ to measure the similarity $\omega$ between each source and target in an adversarial manner. The samples that are closer to the target are selected to distill the source classifier $C^{\prime}$. The prediction of different distilled source classifiers are aggregated based on the domain similarity to obtain the final prediction of the target samples.}
   \label{fig:problem}
\end{figure}

%%%%%%%%% BODY TEXT
%\vspace{-0.3cm}
\section{Introduction}
\label{sec:Introduction}

One key element of the significant success of deep learning algorithms is the availability of large-scale labeled data~\cite{he2016deep}. However, in many practical applications, only limited or even no training data is provided. On the one hand, it is prohibitively labor-intensive and expensive to obtain abundant labeled data. On the other hand, visual data possess variance in nature, which fundamentally limits the scalability and applicability of supervised learning models for handling new scenarios with few labeled examples~\cite{ni2019dual}. In such cases, conventional deep learning approaches suffer from performance decay. Directly transferring the learned models trained on labeled source domains to unlabeled target domains may result in unsatisfying performance, because of the presence of domain shift~\cite{torralba2011unbiased}, which calls for domain adaptation (DA) methods~\cite{bousmalis2016domain,zhao2018emotiongan,hoffman2018cycada}. Unsupervised DA (UDA) addresses such problems by establishing knowledge transfer from a labeled source domain to an unlabeled target domain, and exploring domain-invariant structures and representations to bridge the domain gap~\cite{netzer2011reading}. Both theoretical results~\cite{ben2010theory,gopalan2014unsupervised,tzeng2017adversarial} and algorithms for domain adaptation~\cite{pan2010survey,long2015learning,hoffman2018cycada,zhao2019cycleemotiongan} have been proposed recently.
%, which can be grouped into three categories: the discrepancy-based DA approach [MM-34,49,12,38,1,59]; the adversarial-based approach [MM-29,48,MDAN]; and the reconstruction-based approach [MM-20,58].

\begin{figure*}[!t]
\begin{center}
\centering \includegraphics[width=1.0\linewidth]{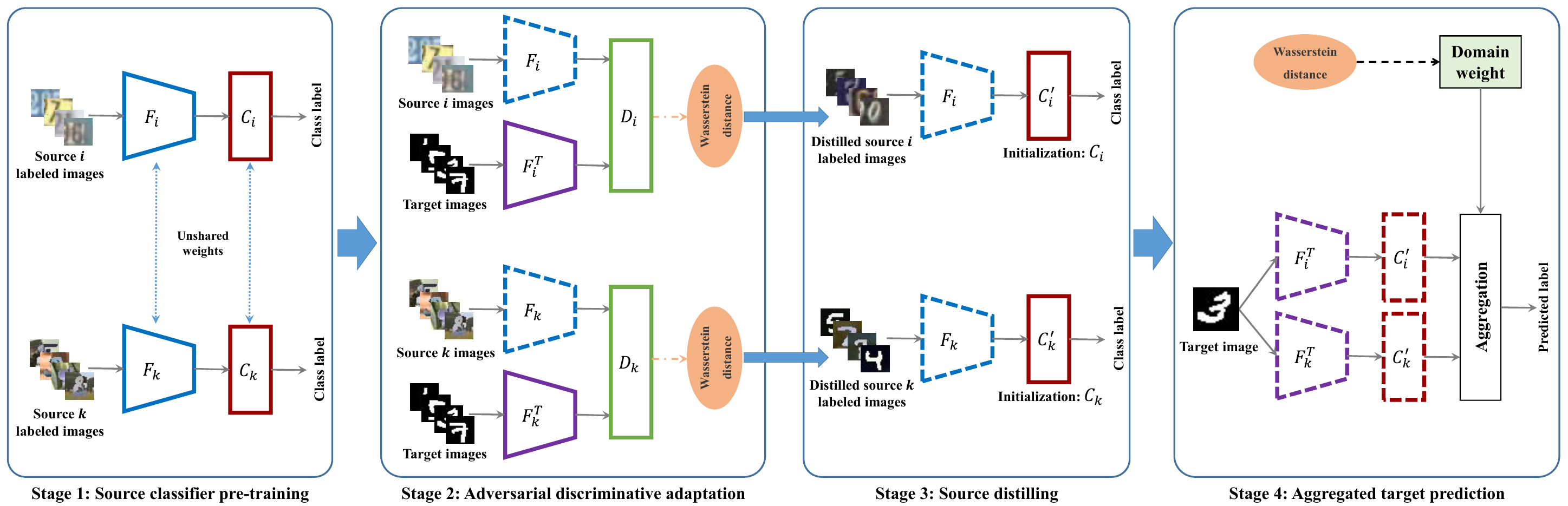}
\caption{The framework of the proposed multi-source distilling domain adaptation (MDDA) network. Dashed rectangles and trapezoids indicate fixed network parameters. $F$, $C$, and $D$ are short for feature extractor, classifier, and domain discriminator, respectively. For simplicity, we just consider the $i$th and $k$th source domains. The Proposed MDDA consists of four stages, as shown from left to right: source classifier pre-training, adversarial discriminative adaptation, source distilling, and aggregated target prediction. Best viewed in color.}
\label{fig:Framework}
\end{center}
\end{figure*}

%The framework of the proposed multi-source distilling domain adaptation (MDDA) network. Dashed rectangles and trapezoids indicate fixed network parameters. $F$, $C$, and $D$ are short for feature extractor, classifier, and domain discriminator, respectively. For simplicity, we just consider the $i$th and $k$th source domains. The Proposed MDDA consists of four stages. \textbf{Stage 1}: We first pre-train a feature extractor (encoder CNN) and classifier for each labeled source domain. \textbf{Stage 2}: We learn a target encoder by adversarial discriminative adaptation for each source such that the estimated Wasserstein distance between encoded target and source features is minimized. \textbf{Stage 3}: We select the source samples that are more similar to the target domain to distill the pre-trained source classifiers. \textbf{Stage 4}: During testing, we map a target image with each learned target encoder to extract features, classify the target image using corresponding source classifier, and aggregate different predictions using domain weight to obtain the final prediction. Best viewed in color.

Though these methods make progress on DA, most of them focus on the single-source setting~\cite{sun2011two,ganin2016domain} and fail to consider a more practical scenario in which there are multiple labeled source domains with different distributions. Naive application of the single-source DA algorithms may lead to suboptimal solution~\cite{shen2017wasserstein}, which calls for effective multi-source domain adaptation (MDA) techniques. Recently, some deep MDA approaches have been proposed~\cite{zhao2018adversarial,xu2018deep,li2018extracting,peng2019moment,zhao2019multi}, but most of them suffer from the following limitations. (1) They sacrifice the discriminative property of the extracted features for the desired task learner in order to learn domain invariant features. (2) They treat the multiple sources equally and fail to consider the different discrepancy among sources and target, as illustrated in Figure~\ref{fig:problem}. Such treatment may lead to suboptimal performance when some sources are very different from the target~\cite{zhao2018adversarial}. (3) They treat different samples from each source equally, without distilling the source data based on the fact that different samples from the same source domain may have different similarities from the target. (4) The adversarial learning based methods suffer from vanishing gradient problem when the domain classifier network can perfectly distinguish target representations from the source ones.

In this paper, we propose a novel multi-source distilling domain adaptation (MDDA) network to address the above challenges by thoroughly exploring the relationships among different sources and the target. As shown in Figure~\ref{fig:Framework}, MDDA can be divided into four stages. (1) We first pre-train the source classifiers separately using the training data from each source. (2) We fix the feature extractor of each source and adversarially map the target into the feature space of each source respectively by minimizing the empirical Wasserstein distance between the source and target~\cite{arjovsky2017wasserstein}, which provides more stable gradients even when the target and source distributions are non-overlap. (3) We select the source training samples that are closer to the target to fine-tune the source classifiers. (4) We build the target predictor by aggregating the source predictions based on the source domain weights, which corresponds to the discrepancy between each source and target. We propose a mechanism to automatically choose a weighting strategy over source domains to emphasize more relevant sources and suppress the irrelevant ones, and aggregate multiple source classifiers based on these weights. With the above four stages, the proposed MDDA can extract features that are both discriminative for the learning task and indiscriminate with respect to the shift among the multiple source and target domains.
%With these four stages, the proposed MDDA achieves two key requirements for domain adaptation: extracting features that are both discriminative for the main learning task on the source domain and indiscriminate with respect to the shift among the multiple-source and target domains.

The main contributions of this paper are summarized as follows: 
\begin{packed_itemize}
\item We propose MDDA to explore the relationships among different sources and target, and achieve more accurate inference on the target by finetuning and aggregating the source classifiers based on these relationships. 
\item Compared to~\cite{xu2018deep}, which symmetrically maps the multiple sources and target into the same space, MDDA learns more discriminative target representations and avoids the oscillation from the simultaneous changing of the multi-source and target distributions by using separate feature extractors that asymmetrically map the target to the feature space of the source in an adversarial manner. Wasserstein distance is used in the adversarial training to achieve more stable gradients even when the target and source distributions are non-overlap. 
\item We propose the source distilling mechanism to select the source training samples that are closer to the target and fine-tune the source classifiers with these samples. 
\item We propose a novel mechanism to automatically choose a weighting strategy over source domains to emphasize more relevant sources and suppress the irrelevant ones, and aggregate the multiple source classifiers based on these weights to build more accurate target predictor. 
\item We extensively evaluate MDDA on the public benchmarks, achieving the state-of-the-art performance and verifying the efficacy of MDDA.
\end{packed_itemize}

%The contributions of our work can be summarized as follows: (1) MDDA learns discriminative representations by using separate feature extractors that map the target data to the same space of each source, respectively, via asymmetric mapping learned through the domain-adversarial loss. Compared to~\cite{xu2018deep}, which symmetrically maps the multiple sources and target into the same space, MDDA learns more discriminative features and avoids the oscillation caused by the simultaneous changing of the multi-source and target distributions in~\cite{xu2018deep}. (2) We propose the source distilling mechanism to select the source training samples that are close to the target and fine-tune the source classifiers using these samples, which utilize more relevant training data and further improve the target performance on the aggregated source classifiers. (3) We propose the pairwise weighting and group weighting mechanisms based on the distance among multiple sources and target to aggregate multiple source classifiers and improve the main learning task accuracy. Such mechanisms explicitly consider the different discrepancy among sources and target and mitigate the detrimental impact of sources that are significantly different from the target. (4) We extensively evaluate MDDA on the public benchmarks, achieving the state-of-the-art performance and verifying the efficacy of MDDA.

%We also interpret the rationale of MDDA by visualizing the extracted features and ablation studies.

\section{Related Work}
\label{sec:RelatedWork}

\noindent\textbf{Single-source UDA} The emphasis of recent single-source UDA (SUDA) methods has shifted to deep learning architectures in an end-to-end fashion. Most deep SUDA methods employ a conjoined architecture with two streams to respectively represent the models for the source domain and the target domain~\cite{zhuo2017deep}. Generally, these methods are trained jointly with a traditional task loss based on the labeled source data and another loss to tackle the domain shift problem, such as discrepancy loss, adversarial loss, reconstruction loss, \emph{etc}.
Discrepancy-based methods explicitly measure the discrepancy between the source and target domains of the two network streams, such as the multiple kernel variant of maximum mean discrepancies~\cite{long2015learning}, correlation alignment (CORAL)~\cite{sun2016return,sun2017correlation,zhuo2017deep}, and contrastive domain discrepancy~\cite{kang2019contrastive}. Adversarial generative models combine the domain discriminative model with a generative component to generate fake source or target data generally based on GAN~\cite{goodfellow2014generative} and its variants, such as CoGAN~\cite{liu2016coupled}, SimGAN~\cite{shrivastava2017learning}, CycleGAN~\cite{zhu2017unpaired,zhao2019cycleemotiongan}, and CyCADA~\cite{hoffman2018cycada}. Adversarial discriminative models usually employ an adversarial objective with respect to a domain discriminator to encourage domain confusion~\cite{ganin2016domain,tzeng2017adversarial,chen2017no,shen2017wasserstein,tsai2018learning,huang2018domain}. Most of these methods suffer from low accuracy when directly applied to the MDA problem.

\noindent\textbf{Multi-source DA} MDA assumes training data are collected from multiple sources~\cite{sun2015survey,zhao2019multi}. There are some theoretical analysis~\cite{ben2010theory,hoffman2018algorithms} to support existing MDA algorithms. The early MDA methods mainly focus on shallow models, including two categories~\cite{sun2015survey}: feature representation approaches~\cite{sun2011two,duan2012exploiting,chattopadhyay2012multisource,duan2012domain} and combination of pre-learned classifiers \cite{xu2012multi,sun2013bayesian}. Some novel shallow MDA methods aim to deal with special cases, such as incomplete MDA~\cite{ding2018incomplete} and target shift~\cite{redko2019optimal}.

Recently, some representative deep learning based MDA methods are proposed, such as multisource domain adversarial network (MDAN)~\cite{zhao2018adversarial}, deep cocktail network (DCTN)~\cite{xu2018deep}, and moment matching network (MMN)~\cite{peng2019moment}. All these MDA methods employ a shared feature extractor network to symmetrically map the multiple sources and target into the same space. %, without considering the conditional shift problem.
For each source-target pair in MDAN and DCTN, a discriminator is trained to distinguish the source and target features. MDAN concatenates all extracted source features and labels into one domain to train a single task classifier, while DCTN trains a classifier for each source domain and combines the predictions of different classifiers for a target image using perplexity scores as weights. MMN transfers the learned knowledge from multiple sources to the target by dynamically aligning moments of their feature distributions. The final prediction of a target image is averaged uniformly based on the classifiers from different source domains.
Different from these works, we employ an unshared feature extractor to obtain the feature representation for each source, match the target feature to each source feature space asymmetrically, distill the pre-trained classifiers with selected representative samples, and combine the predictions of different classifiers using a novel weighting strategy.

\section{Problem Definition}
\label{sec:ProblemDefinition}
Suppose we have $M$ source domains $S_1,S_2,\cdots,S_M$ and one target domain $T$. In unsupervised domain adaptation (UDA) scenario, $S_1,S_2,\cdots,S_M$ are labeled and $T$ is fully unlabled. For the $i$th source domain $S_i$, the observed images and corresponding labels drawn from the source distribution $p_i(\textbf{x},y)$ are $\textbf{X}_i=\{\textbf{x}_i^j\}_{j=1}^{N_i}$ and $Y_i=\{y_i^j\}_{j=1}^{N_i}$, where $N_i$ is the number of source images. The target images drawn from the target distribution $p_T(\textbf{x},y)$ are $\textbf{X}_T=\{\textbf{x}_T^j\}_{j=1}^{N_T}$ without label observation, where $N_T$ is the number of target images.
%Our method is based on two basic hypotheses of variation between the source and target domains: (1) covariate shift, $p_1(y|\textbf{x})=\cdots=p_M(y|\textbf{x})=p_T(y|\textbf{x})$ for all $\textbf{x}$, but $p_1(\textbf{x})\neq\cdots \neq p_M(\textbf{x})\neq p_T(\textbf{x})$; (2) concept drift, $p_1(\textbf{x},y)\neq\cdots \neq p_M(\textbf{x},y)\neq p_T(\textbf{x},y)$.
%fundamental hypotheses
Unless otherwise specified, we assume (1) homogeneity, \textit{i.e.} $\textbf{x}_i^j\in \mathds{R}^{d}, \textbf{x}_T^j\in \mathds{R}^{d}$, which indicates that the data from different domains are observed in the same feature space but exhibit different distributions; (2) closed set, \textit{i.e.} $y_i^j\in \mathcal{Y}, y_T^j\in \mathcal{Y}$, where $\mathcal{Y}$ is the class label space, indicating that all the domains share their categories. Our goal is to learn an adaptation model that can correctly predict a sample from the target domain based on $\{(\textbf{X}_i,Y_i)\}_{i=1}^{M}$ and $\{\textbf{X}_T\}$. Please note that our method can be easily extended to tackle heterogeneous DA~\cite{li2014learning,hubert2016learning} by changing the network structure of the target feature extractor, open set DA~\cite{panareda2017open} by adding an ``unknown'' class, or category shift DA~\cite{xu2018deep} by reweighing the predictions of only those domains that contain the specified category. We will investigate such study in our future work.

\section{Multi-source Distilling Domain Adaptation}
\label{sec:MDDA}
In this section, we introduce the proposed multi-source distilling domain adaptation (MDDA) network. MDDA is a novel approach to overcome the limitations of existing methods for multiple source domain adaptation by thoroughly exploring the relationships among different sources and the target. It achieves more accurate inference on the target by finetuning and aggregating the source classifiers based these relationships. As shown in Figure~\ref{fig:Framework}, MDDA can be divided into four stages. We first pre-train the source classifiers separately with the training data from each source. Then, we fix the feature extractor of each source and map the target into the feature space of each source adversarially by minimizing the estimated Wasserstein distance between the source and target. MDDA learns more discriminative target representations and avoids the oscillation from the simultaneous changing of the multi-source and target distributions by using separate feature extractors that asymmetrically map the target to the feature space of the source in an adversarial manner. In the third stage, the source samples closer to the target are selected to fine-tune the source classifiers. Finally, we build the target predictor by aggregating the source predictions based on the discrepancy between each source and target. We propose a novel mechanism to automatically choose a weighting strategy over source domains to emphasize more relevant sources and suppress the irrelevant ones. With the above four stages, MDDA extracts features that are both discriminative for the learning task and indiscriminate with respect to the shift among the multiple source and target domains. We will explain each stage in the following subsections.

%MDDA is novel from the following aspects: (1) It learns separate encoder mappings of the target image to each source feature space such that a domain discriminator cannot distinguish the encoded target features from the source features. Such adversarial process learns both domain invariant and task discriminative features. The exploitation of WGAN enables MDDA to deal with the conditional shift challenge. (2) The source distilling mechanism selects source training samples that are close to the target and fine-tunes source classifiers using these samples, which utilize more relevant training data and improve the performance of the aggregated source classifiers on the target. (3) The classification of a target image can be obtained by combining the predictions of the source classifiers, which are reweighted based on the distance among multiple sources and target. As shown in Figure~\ref{fig:Framework}, MDDA is composed of four components: source classifier pre-training, adversarial discriminative adaptation, source distilling, and aggregated target prediction.%, which will be explained in the following subsections.

\subsection{Source Classifier Pre-training}
\label{ssec:pre-training}

To extract more task discriminative features and learn accurate classifiers, we pre-train a feature extractor $F_i$ and classifier $C_i$ for each labeled source domain $S_i$ with unshared weights between different domains. Take the $N$-class classification task as an example, $F_i$ and $C_i$ are optimized by minimizing the following cross-entropy loss:
%\begin{equation}
%\small
%\mathcal{L}_{cls}(F_i,C_i)=\mathbb{E}_{(\textbf{x}_i,y_i)\sim p_i}\sum_{n=1}^{N}\mathds{1}_{[n=y_i]}\log(\sigma(C_i(F_i(\textbf{x}_i)))),
%\label{equ:f_loss}
%\end{equation}
\begin{equation}
\small
\begin{aligned}
&\mathcal{L}_{cls}(F_i,C_i)=\\
&\ \ \ \ \ \ -\mathbb{E}_{(\textbf{x}_i,y_i)\sim p_i}\sum_{n=1}^{N}\mathds{1}_{[n=y_i]}\log(\sigma(C_i(F_i(\textbf{x}_i)))),\\
\end{aligned}
\label{equ:cls_loss}
\end{equation}
where $\sigma$ is the softmax function, and $\mathds{1}$ is an indicator function. Comparing with a shared feature extractor network to extract domain-invariant features among different source domains~\cite{zhao2018adversarial,xu2018deep}, the unshared feature extractor network can obtain the discriminative feature representations and accurate classifiers for each source domain. When aggregating the multiple predictions based on the source classifier and matched target features in the later stage, the final target prediction would be better boosted.

\subsection{Adversarial Discriminative Adaptation}
\label{ssec:Training}
%After the pre-training stage, we learn separate target encoder $F_i^T$ to map the target feature into the same space of source $S_i$. Meanwhile, a discriminator $D_i$ is trained to distinguish whether the feature of an image originates from the target $T$ or the source $S_i$. We employ an adversarial loss to optimize $F_i^T$ and $D_i$. In other words, $D_i$ aims to maximize the probability of correctly classifying the encoded target features from $F_i^T$ and the encoded source feature from pre-trained $F_i$, while $F_i^T$ tries to maximize the probability of $D_i$ making a mistake. Similar to GAN~\cite{goodfellow2014generative}, we model this as a two-player minimax game. $D_i$ is obtained by maximizing
After the pre-training stage, we learn separate target encoder $F_i^T$ to map the target feature into the same space of source $S_i$. A discriminator $D_i$ is trained adversarially to maximize the Wasserstein distance of correctly classifying the encoded target features from $F_i^T$ and the encoded source feature from pre-trained $F_i$, while $F_i^T$ tries to maximize the probability of $D_i$ making a mistake, \textit{i.e.} minimizing the Wasserstein distance. Similar to GAN~\cite{goodfellow2014generative}, we model this as a two-player minimax game. Following~\cite{arjovsky2017wasserstein}, we suppose the discriminators $\{D_i\}$ are all 1-Lipschitz and then we can optimize $D_i$ by maximizing the Wasserstein distance
\begin{equation}
\small
\mathcal{L}_{wd_D}(D_i)=\mathbb{E}_{\textbf{x}_i\sim p_i} D_i(F_i(\textbf{x}_i))-\mathbb{E}_{\textbf{x}_T\sim p_T} [D_i(F_i^T(\textbf{x}_T))],
\label{equ:d_loss}
\end{equation}
while $F_i^T$ is obtained by minimizing
\begin{equation}
\mathcal{L}_{wd_F}(F_i^T)=-\mathbb{E}_{\textbf{x}_T\sim p_T} D_i(F_i^T(\textbf{x}_T)).
\label{equ:f_loss}
\end{equation}
In this way, the target encoder $F_i^T$ tries to confuse the discriminator $D_i$ by minimizing the Wasserstein distance between the encoded target features as the source ones.

To enforce the Lipschitz constraint~\cite{goodfellow2014generative}, we add a gradient penalty for the parameters of each discriminator $D_i$ as in~\cite{gulrajani2017improved}
\begin{equation}
\mathcal{L}_{grad}(D_i)=(\|\nabla_{\hat{\textbf{x}}}D_i(\hat{\textbf{x}})\|_2-1)^2,
\label{equ:grad_loss}
\end{equation}
where $\hat{\textbf{x}}$ is a feature set that contains not only the source and target features but also the random points along the straight line between source and target feature pairs~\cite{gulrajani2017improved}. $D_i$ can then be optimized by
\begin{equation}
\max_{D_i}\mathcal{L}_{wd_D}(D_i)-\alpha\mathcal{L}_{grad}(D_i),
\label{equ:d_loss_new}
\end{equation}
where $\alpha$ is a balancing coefficient, the value of which can be empirically set.

\subsection{Source Distilling}
\label{ssec:fine-tuning}

We further dig into each source domain to select the source training samples that are closer to the target based on the estimated Wasserstein distance to fine-tune the source classifiers. Such source distilling mechanism utilizes more relevant training data and further improves the target performance on the aggregated source classifiers. We select the source samples based on the estimated Wasserstein distance, since it can represent the divergence between source data and target data. For each source sample $x_i^j$ in the $i$th source domain, we calculate the Wasserstein distance between each source sample and target domain:
\begin{equation}
\small
\tau_{i}^{j} = || D_i(F_i(x_j)) - \frac{1}{N_T}\sum_{k=1}^{N_T}D_i(F_i^T(x_k)) ||.
\label{equ:wdistance}
\end{equation}

For each source sample $x_i^j$, $\tau_i^j$ reflects the its distance to the target domain. The smaller the $\tau_i^j$ value is, the closer it is to the target domain. Therefore, in each source domain $X_i$, we select $\frac{N_i}{2}$ of the source data $\hat{p}_i$ = $\{\hat{x}_i^j, \hat{y}_i^j\}_{j=1}^{N_i}$ whose $\tau_i^j$ is larger than the left ones.
With these selected source data, we finetune $C_i$ by minimizing the following objective:
\begin{equation}
\small
\begin{aligned}
&\mathcal{L}_{distill}(C_i)=\\
&\ \ \ \ \ \ -\mathbb{E}_{(\hat{\textbf{x}}_i,\hat{y}_i)\sim p_i}\sum_{n=1}^{N}\mathds{1}_{[n=\hat{y}_i]}\log(\sigma(C_i(F_i(\hat{\textbf{x}}_i)))),\\
\end{aligned}
\label{equ:distill}
\end{equation}

\subsection{Aggregated Target Prediction}
\label{ssec:Testing}
In the testing stage, the goal is to accurately classify a given target image $\textbf{x}_T$. Corresponding to each source domain, we extract the features $F_i^T(\textbf{x}_T)$ of the target image based on the learned target encoder from stage 2, and obtain source-specific prediction $C_i^{\prime}(F_i^T(\textbf{x}_T))$ using the distilled source classifier. Next, we combine the different predictions from each source classifier to obtain the final prediction:
\begin{equation}
\small
Result(\textbf{x}_T)=\sum_{i=1}^{N}\omega_{i}C_i^{\prime}(F_i^T(\textbf{x}_T)).
\label{equ:final_label}
\end{equation}
The key problem here is how to select the weights $\omega_{i}$ for the predictions from different source classifiers. We design a novel weighting strategy based on the discrepancy between each source and target to emphasize more relevant sources and suppress the irrelevant ones.
We assume after training in stage 2, the estimated Wasserstein distance $L_{wd_{D_i}}$ between each source $S_i$ and target $T$ subordinates to a standard Gaussian Distribution $\mathcal{N}(0,1)$. Therefore, the weight of each domain can be computed by the following equation
\begin{equation}
\small
\omega_{i} = e^{\frac{-L_{wd_{D_i}}^2}{2}}.
\label{equ:w1weight}
\end{equation}

\begin{table}[!t]
\centering\scriptsize%\small
%\caption{Classification accuracy (\%) on Digits-five dataset for multi-source unsupervised domain adaptation. The best method trained on the source domains is emphasized in bold and red. Our method achieves accuracy, significantly outperforming the state-of-the-arts.}
\caption{Classification accuracy (\%) on Digits-five dataset for multi-source unsupervised domain adaptation. The best method is emphasized in bold. Our method achieves 88.1\% accuracy, significantly outperforming the state-of-the-art approaches.}
\resizebox{\linewidth}{!}{%
\begin{tabular}
{c | c | c | c | c | c | c | c}
\hline
Standards & Models & mm & mt & up & sv & sy & Avg\\
\hline
\multirow{2}{1.3cm}{\centering Source-only}   &  Combined & 63.7  &  92.3 & 87.2  &  66.3  &  84.8  &  78.9\\
 & Single-best  & 59.2  & 97.2  & 84.7  &  77.7 & 85.2 & 80.8  \\
\hline
\multirow{4}{1.3cm}{\centering Single-best DA}   & DAN~\shortcite{long2015learning}  & 63.8  &  96.3 & \textbf{94.2}  & 62.5 & 85.4  &  80.4\\
   & CORAL~\shortcite{sun2016return}  & 62.5  & 97.2  & 93.5  & 64.4 & 82.8  & 80.1 \\
   & DANN~\shortcite{ganin2016domain}  & 71.3  & 97.6  & 92.3  & 63.5  &  85.3 & 82.0 \\
   & ADDA~\shortcite{tzeng2017adversarial}  & 71.6  & 97.9  & 92.8  & 75.5 & 86.5  & 84.9 \\
%   & ADDA WGAN~\shortcite{tzeng2017adversarial}  & 88.9  & 98.8  & 91.6  & 79.5 & 89.6  & 89.7 \\
\hline
\multirow{3}{1.3cm}{\centering Source-combined DA}   & DAN~\shortcite{long2015learning}  &  67.9 & 97.5  & 93.5  &  67.8 & 86.9  & 82.7 \\
   & DANN~\shortcite{ganin2016domain}  & 70.8  & 97.9  &  93.5 &  68.5 & 87.4  & 83.6 \\
   & ADDA~\shortcite{tzeng2017adversarial}  & 72.3  & 97.9  & 93.1  &  75.0 & 86.7  &  85.0 \\
%   & ADDA (ours) & 58.5  & 91.1  & 89.5  &  70.8 & 77.8  &  77.5 \\
\hline
\multirow{3}{1.3cm}{\centering Multi-source DA}   & DCTN~\shortcite{xu2018deep}  & 70.5  & 96.2  & 92.8  & 77.6  & 86.8  &  84.8 \\
   & MDAN~\shortcite{zhao2018adversarial}  & 69.5  & 98.0  &  92.5  & 69.2  & 87.4  & 83.3 \\
   & MDDA (ours)  & \textbf{78.6}  & \textbf{98.8}  &  93.9  &  \textbf{79.3} &  \textbf{89.7} &  \textbf{88.1}\\
\hline
%Oracle   & AlexNet  &   &   &   &  &  & \\
%\hline
\end{tabular}
}
\label{tab:DigitsFive}
\end{table}

\section{Experiments}
\label{sec:Experiments}

We evaluate the proposed MDDA model on multi-source domain adaptation task in visual classification applications, including digit recognition and object classification.% More implementation details are included in the supplementary material.
%In this section, we evaluate the proposed MDDA model on multi-source domain adaptation task in visual classification applications, including digit recognition and object classification. We first introduce the employed benchmarks and compared baselines. And then we report and analyze the major results together with some empirical analysis. More implementation details and results are included in the supplementary material.

% \subsection{Benchmarks}
% \label{ssec:Benchmarks}
\subsection{Experimental Settings}
\label{ssec:Settings}

\subsubsection{Benchmarks}
\label{sssec:Benchmarks}

Digits-five includes 5 digit image datasets sampled from different domains, including \emph{handwritten} \textbf{mt} (MNIST) \cite{lecun1998gradient}, \emph{combined} \textbf{mm} (MNIST-M) \cite{ganin2015unsupervised}, \emph{street image} \textbf{sv} (SVHN) \cite{netzer2011reading}, \emph{synthetic} \textbf{sy} (Synthetic Digits) \cite{ganin2015unsupervised}, and \emph{handwritten} \textbf{up} (USPS) \cite{hull1994database}. Following~\cite{xu2018deep,peng2019moment}, we sample 25,000 images for training and 9,000 for testing in \textbf{mt}, \textbf{mm}, \textbf{sv}, \textbf{sy}, and select the entire 9,298 images in \textbf{up} as a domain.

Office-31~\cite{saenko2010adapting} contains 4,110 images within 31 categories, which are collected from office environment in 3 image domains: \textbf{A} (Amazon) downloaded from amazon.com, \textbf{W} (Webcam) and \textbf{D} (DSLR) taken by web camera and digital SLR camera, respectively.

\subsubsection{Baselines}
\label{ssec:Baselines}

To compare MDDA with the state-of-the-art approaches for MDA, we select the following methods as baselines. \textbf{(1) Source-only}, \textit{i.e.} train on the source domains and test on the target domain directly. We can view this as a lower bound of DA. \textbf{(2) Single-source DA}, perform multi-source DA via single-source DA, including conventional models, \textit{i.e.} TCA~\cite{pan2011domain} and GFK~\cite{gong2012geodesic}, and deep methods, \textit{i.e.} DDC~\cite{tzeng2015simultaneous}, DRCN~\cite{ghifary2016deep}, RevGrad~\cite{ganin2015unsupervised}, DAN~\cite{long2015learning}, RTN~\cite{long2016unsupervised}, CORAL~\cite{sun2016return}, DANN~\cite{ganin2016domain}, and ADDA~\cite{tzeng2017adversarial}. \textbf{(3) Multi-source DA}, extend some single-source DA method to multi-source settings, including DCTN~\cite{xu2018deep} and MDAN~\cite{zhao2018adversarial}.

\begin{table}[!t]
\centering\scriptsize%\small
\caption{Classification accuracy (\%) on Office31 dataset for multi-source unsupervised domain adaptation. The best method is emphasized in bold. Our method achieves 84.2\% accuracy, achieving the state-of-the-art performances.}%significantly outperforming the state-of-the-art approaches.
%\resizebox{\linewidth}{!}{%
\begin{tabular}
{c | c | c | c | c | c }
\hline
%\multirow{2}{*}{Standards} & \multirow{2}{*}{Models} & A,W & A,D & D,W & \multirow{2}{*}{\ Avg\ }\\
%   &   &  $\rightarrow$D & $\rightarrow$W  & $\rightarrow$A  & \\
Standards & Models & D & W & A & Avg\\
\hline
\multirow{2}{1.3cm}{\centering Source-only}   &  Combined & 97.1  & 92.0  & 51.6  & 80.2 \\
 & Single-best  & 99.0  & 95.3  & 50.2  &  81.5 \\
\hline
\multirow{8}{1.3cm}{\centering Single-best DA}   & TCA~\shortcite{pan2011domain}  &  95.2 &  93.2 &  51.6 & 80.0 \\
& GFK~\shortcite{gong2012geodesic}  & 95.0  &  95.6 &  52.4 &  81.0 \\
& DDC~\shortcite{tzeng2015simultaneous}  & 98.5  & 95.0  &  52.2 & 81.9 \\
& DRCN~\shortcite{ghifary2016deep}  & 99.0  & 96.4  &  56.0 & 83.8 \\
& RevGrad~\shortcite{ganin2015unsupervised}  & 99.2  & 96.4  &  53.4 & 83.0 \\
& DAN~\shortcite{long2015learning}  & 99.0  & 96.0  & 54.0  & 83.0 \\
& RTN~\shortcite{long2016unsupervised}  & \textbf{99.6}  & 96.8  & 51.0  & 82.5 \\
& ADDA~\shortcite{tzeng2017adversarial}  & 99.4  & 95.3  & 54.6  & 83.1 \\
%& ADDA(ours) & 99.2  & 96.0  & 54.6  & 83.2 \\
\hline
\multirow{3}{1.3cm}{\centering Source-combined DA}   & RevGrad~\shortcite{ganin2015unsupervised}  & 98.8  & 96.2  &  54.6 & 83.2 \\
   & DAN~\shortcite{long2015learning}  & 98.8  & 96.2  & 54.9  &  83.3 \\
   & ADDA~\shortcite{tzeng2017adversarial}  & 99.2  &  96.0 & 55.9  & 83.7  \\
%   & ADDA(ours) & 98.2  & 96.9  & 52.6  & 82.6 \\
\hline
\multirow{3}{1.3cm}{\centering Multi-source DA}   & DCTN~\shortcite{xu2018deep}  & \textbf{99.6}  &  96.9 & 54.9  & 83.8  \\
   & MDAN~\shortcite{zhao2018adversarial}  & 99.2  & 95.4  &  55.2 &  83.3  \\
   & MDDA (ours)  & 99.2  & \textbf{97.1}  & \textbf{56.2}  &  \textbf{84.2}  \\
\hline
%Oracle   & AlexNet  &   &   &   &   \\
%\hline
\end{tabular}
%}
\label{tab:Office31}
\end{table}

%To compare the proposed MDDA model with the state-of-the-art approaches for multi-source UDA, we select the following methods as baselines. \textbf{(1) Source-only}, \textit{i.e.} train on the source domains and test on the target domain directly. We can view this as a lower bound of DA. \textbf{(2) Single-source DA}, perform multi-source DA via single-source DA, including conventional models, \textit{i.e.} Transfer Component Analysis (TCA)~\cite{pan2011domain} and Geodesic Flow Kernel (GFK)~\cite{gong2012geodesic}, and deep methods, \textit{i.e.} Deep Domain Confusion (DDC)~\cite{tzeng2015simultaneous}, Deep Reconstruction-classification Networks (DRCN)~\cite{ghifary2016deep}, Reversed Gradient (RevGrad)~\cite{ganin2015unsupervised}, Domain Adaptation Network (DAN)~\cite{long2015learning}, Residual Transfer Network (RTN)~\cite{long2016unsupervised}, Correlation Alignment (CORAL)~\cite{sun2016return}, Domain Adversarial Neural Network (DANN)~\cite{ganin2016domain}, and Adversarial Discriminative Domain Adaptation (ADDA)~\cite{tzeng2017adversarial}. \textbf{(3) Multi-source DA}, extend some single-source DA method to multi-source settings, including Deep Cocktail Network (DCTN)~\cite{xu2018deep} and Multisource Domain Adversarial Network (MDAN)~\cite{zhao2018adversarial}.

%and \textbf{(4) Oracle}, train on the target domain, which can be viewed as an upper bound of DA.

For the source-only and single-source DA standards, we employ two strategies: (1) source-combined, \textit{i.e.} all source domains are combined into a traditional single source; (2) single-best, \textit{i.e.} performing adaptation on each single source and selecting the best adaptation result in the target test set.

\subsubsection{Implementation Details}
\label{sssec:Details}
In Digits-five experiments, we use three convlutional layers and two fully connected layers as encoder and one fully connected layer as classifier. In Office-31 experiments, we use Alexnet as our backbone. The last layer is used as classifier and the other layers are used as encoder. Following~\cite{gulrajani2017improved}, we set $\alpha$ in Eq.~(\ref{equ:d_loss_new}) to 10.
%More implementation details can be found in the supplementary material.

%\begin{itemize}
% \item Source-only, i.e. train on the source domains and test on the target domain directly. We can view this as a lower bound of DA.
% \item Single-source DA, performs multi-source DA via single-source DA, including conventional models, i.e. Transfer Component Analysis (TCA)~\cite{pan2011domain} and Geodesic Flow Kernel (GFK)~\cite{gong2012geodesic}, and deep methods, i.e. Deep Domain Confusion (DDC)~\cite{tzeng2015simultaneous}, Deep Reconstruction-classification Networks (DRCN)~\cite{ghifary2016deep}, Reversed Gradient (RevGrad)~\cite{ganin2015unsupervised}, Domain Adaptation Network (DAN)~\cite{long2015learning}, Residual Transfer Network (RTN)~\cite{long2016unsupervised}, Correlation Alignment (CORAL)~\cite{sun2016return}, Domain Adversarial Neural Network (DANN)~\cite{ganin2016domain}, and Adversarial Discriminative Domain Adaptation (ADDA)~\cite{tzeng2017adversarial}.
% \item Multi-source DA, extends some single-source DA method to multi-source settings, including Deep Cocktail Network (DCTN)~\cite{xu2018deep} and Multisource Domain Adversarial Network (MDAN)~\cite{zhao2018adversarial}.
% \item Oracle, train on the target domain, which can be viewed as an upper bound of DA.
%\end{itemize}

\begin{figure*}[!t]
\begin{center}
\centering \includegraphics[width=0.8\linewidth]{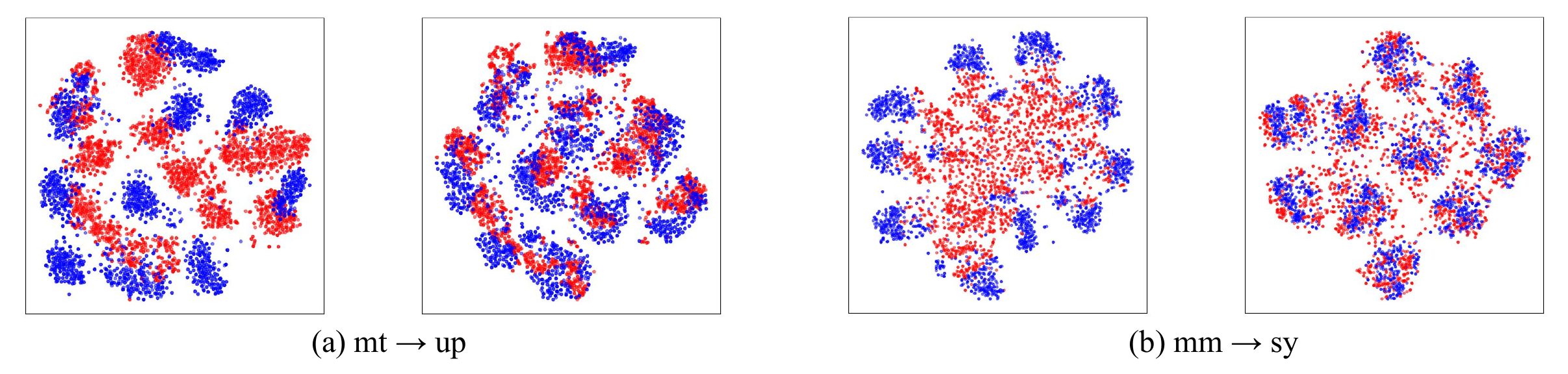}
\caption{The t-SNE~\cite{maaten2008visualizing} visualization of of the digit-5 dataset features for (a) mt$\rightarrow$up and (b) mm$\rightarrow$sy. In each pair, the features are extracted using the last layer of source domain encoder from the samples of source and target domain in the first image, and the target domain features are extracted using the the last layer of adapted encoder in the second one.}
\label{fig:Visualization}
\end{center}
\end{figure*}

\subsection{Comparison with the State-of-the-art}
\label{ssec:Comparison}

The performance comparisons between MDDA and the state-of-the-art approaches as measured by classification accuracy are shown in Table~\ref{tab:DigitsFive} and Table~\ref{tab:Office31} on Digits-five and Office-31 datsets, respectively. From the results, we have the following observations.

\textbf{(1)} The source-only method \textit{i.e.} directly transferring the models trained on the source domains to the target domain performs the worst in most adaptation settings. Due to the presence of domain shift, the joint probability distributions of observed images and class labels greatly differ in the source and target domains. This results in the model's low transferability from the source domains to the target domain. Further, even with more training samples, the Combined setting does not guarantee to perform better than the Single-best one. This is because domain shift also exists across different source domains, which may confuse the classifier. For example, if one source domain and the target is very similar, such as sv and sy, and the other source domains are quite different, simple combination would enlarge the domain shift between the Single-best and the target. This observation demonstrates the necessity of designing DA algorithms to address the domain shift problem.
%, e.g. two samples from different domains may have similar extracted features but different labels.

\textbf{(2)} Almost all adaptation methods outperform the source-only methods, demonstrating the effectiveness of DA in image classification. Comparing the Single-best DA and Source-combined DA, it is clear that on average the Source-combined DA performs better, which is different from the source-only scenario. This is because after adaptation, domain-invariant representations are learned for the samples of different domains. Therefore, the Source-combined DA works better with the help of more training data.

\textbf{(3)} Generally, multi-source DA performs better than other adaptation standards. This is more clear when comparing the methods that employ similar adaptation architectures, such as our MDDA vs. ADDA~\cite{tzeng2017adversarial} and MDAN~\cite{zhao2018adversarial} vs. DANN~\cite{ganin2016domain}. Not only the domain shift between the sources and the target, but also the shift across the different source domains is bridged in multi-source DA, which boosts the adaptation by exploring the complementarity of different sources.

\begin{table}[!t]
\centering\scriptsize%\small
\caption{Ablation study of different weighting strategies in the proposed MDDA model on Digits-five dataset for multi-source unsupervised domain adaptation.}
%\resizebox{\linewidth}{!}{%
\begin{tabular}
{c | c | c | c | c | c | c}
\hline
%\multirow{2}{*}{Weighting} & \tiny{mt,up,sv,} & \tiny{mm,up,sv,} & \tiny{mm,mt,sv,} & \tiny{mm,mt,up,} & \tiny{mm,mt,up,} & \multirow{2}{*}{Avg}\\
%&  \tiny{sy$\rightarrow$mm} & \tiny{sy$\rightarrow$mt}  & \tiny{sy$\rightarrow$up}  & \tiny{sy$\rightarrow$sv} &  \tiny{sv$\rightarrow$sy}  & \\
Weighting & mm & mt & up & sv & sy & Avg\\
\hline
Uniform    & 74.3  & 95.8  & 93.7  & 64.2  & 79.3  &  81.5  \\
Ours    & 78.6  & 98.8  &  93.9  &  79.3 &  89.7 &  88.1\\
\hline
\end{tabular}
%}
\label{tab:DigitsFiveAblation}
\end{table}

\begin{table}[!t]
\centering\scriptsize%\small
\caption{Ablation study of different weighting strategies in the proposed MDDA model on Office31 dataset for multi-source unsupervised domain adaptation.}
\begin{tabular}
{c | c | c | c | c }
\hline
%Weighting & A,W$\rightarrow$D & A,D$\rightarrow$W & D,W$\rightarrow$A & Avg\\
Weighting & D & W & A & Avg\\
\hline
Uniform    & 98.4  & 95.2  & 55.7  & 83.1  \\
Ours   & 99.2  & 97.1  & 56.2  &  84.2  \\
\hline
\end{tabular}
\label{tab:Office31Ablation}
\end{table}

%which benefits

\textbf{(4)} The proposed MDDA model performs better than state-of-the-art multi-source methods in most cases.
On one hand, the performance improvements of MDDA over the best Source-combined method are 3.1\% and 0.5\% on Digits-five and Office-31 datsets, respectively. On the other hand, the proposed MDDA method achieves 3.3\%, 4.8\% and 0.4\%, 0.9\% performance improvements as compared to DCTN~\cite{xu2018deep} and MDAN~\cite{zhao2018adversarial} on Digits-five and Office-Home datsets, respectively.
These results demonstrate that the proposed MDDA model can achieve superior performance relative to state-of-the-art approaches.
The performance improvements benefit from the advantages of MDDA. First, the unshared weights enable to learn the best feature extractor and classifiers for each source domain, which would boost the performance when aggregation. Second, a novel weighting strategy based on the Wasserstein distance can better emphasize the domains that are more closer to the target. Finally, for each source domain, selective samples are distilled to fine-tune the source classifier, which also adapt better to the target features.

\subsection{Interpretability and Ablation Study}
\label{ssec:EmpiricalAnalysis}

%\textbf{Feature Visualization} 

\subsubsection{Feature Visualization} 

To show the adaptation ability of the proposed MDDA model, we visualize the features before and after adversarial adaptation with t-SNE embedding~\cite{maaten2008visualizing} in tasks: mt$\rightarrow$up and mm$\rightarrow$ sy. As illustrated in Figure~\ref{fig:Visualization}, we have two observations: (1) target features become more dense while using adversarial adaptation; (2) target domain fits source domain more tightly after the adversarial adaptation, which demonstrates that MDDA can align the distributions between the source and target domains.

%To show the adaptation ability of the proposed MDDA model, we visualize the features before and after adversarial adaptation with t-SNE embedding~\cite{maaten2008visualizing} in task: synth -> mnistm. As illustrated in Figure~\ref{fig:Visualization}, we have three observations: (1) after adversarial adaption, target features become more dense; (2) and target domain fits source domain more tightly after the adversarial adaption (3) In the right one, the feature distributions are a few different between the synth and mnistm, since the original images are very different in the two domains.

%\textbf{Model interpretability}.

\begin{table}[!t]
\centering\scriptsize%\small
\caption{Ablation study of whether distilling the source classifiers in the proposed MDDA model on Digits-five dataset for multi-source unsupervised domain adaptation.}
%\resizebox{\linewidth}{!}{%
\begin{tabular}
{c | c | c | c | c | c | c}
\hline
%& \tiny{mt,up,sv,} & \tiny{mm,up,sv,} & \tiny{mm,mt,sv,} & \tiny{mm,mt,up,} & \tiny{mm,mt,up,} & \multirow{2}{*}{Avg}\\
%&  \tiny{sy$\rightarrow$mm} & \tiny{sy$\rightarrow$mt}  & \tiny{sy$\rightarrow$up}  & \tiny{sy$\rightarrow$sv} &  \tiny{sv$\rightarrow$sy}  & \\
 & mm & mt & up & sv & sy & Avg\\
\hline
w/o  & 78.4  & 98.8  & 93.2  & 79.1  & 89.6  &  87.8  \\
w     & 78.6  & 98.8  &  93.9  &  79.3 &  89.7 &  88.1\\
\hline
\end{tabular}
%}
\label{tab:DigitsFivefine-tune}
\end{table}

\begin{table}[!t]
\centering\scriptsize%\small
\caption{Ablation study of whether distilling the source classifiers in the proposed MDDA model on Office31 dataset for multi-source unsupervised domain adaptation.}
\begin{tabular}
{c | c | c | c | c }
\hline
% & A,W$\rightarrow$D & A,D$\rightarrow$W & D,W$\rightarrow$A & Avg\\
& D & W & A & Avg\\
\hline
w/o   & 99.2  & 96.0  & 55.8  & 83.7  \\
w   & 99.2  & 97.1  & 56.2  &  84.2  \\
\hline
\end{tabular}
\label{tab:Office31fine-tune}
\end{table}

\subsubsection{Ablation Study} 

The proposed MDDA model contains two major components: source distilling for fine-tuning the source classifiers and a novel weighting strategy for aggregating target prediction. We conduct ablation study to further verify their effectiveness by changing one component while fixing the other.

We compare the proposed weighting strategy with one straightforward baseline: uniform weight. The results on Digits-five and Office31 datasets are shown in Table~\ref{tab:DigitsFiveAblation} and Table~\ref{tab:Office31Ablation}, respectively. From the results, we can observe that the proposed weighting strategy outperforms the uniform weight. This is reasonable because the uniform weight does not reveal the importance of different sources, which might have different similarities to the target. By considering the relative similarity of different sources to the target based on the Wasserstein distance, the proposed MDDA achieves 6.6\% and 1.1\% improvements on Digits-five and Office31 datasets, respectively. These observations demonstrate the effectiveness of the proposed weighting strategy.

\begin{table}[!t]
\centering\scriptsize%\small
\caption{An example of detailed distilling result from each source to the target sy on Digits-five dataset.}
%\resizebox{\linewidth}{!}{%
\begin{tabular}
{c | c | c | c | c | c}
\hline
%\multirow{2}{*}{Weighting} & \tiny{mt,up,sv,} & \tiny{mm,up,sv,} & \tiny{mm,mt,sv,} & \tiny{mm,mt,up,} & \tiny{mm,mt,up,} & \multirow{2}{*}{Avg}\\
%&  \tiny{sy$\rightarrow$mm} & \tiny{sy$\rightarrow$mt}  & \tiny{sy$\rightarrow$up}  & \tiny{sy$\rightarrow$sv} &  \tiny{sv$\rightarrow$sy}  & \\
Source & mt & mm & sv & up & Avg\\
\hline
w/o    & 52.0  & 70.8  & 89.4  & 38.6  &  62.7  \\
w    & 54.5  & 71.0  &  89.5  &  40.8 &  64.0 \\
% w/o    & 52.00  & 70.76  & 89.44  & 38.62  &  62.71  \\
% w    & 54.51  & 70.96  &  89.45  &  40.78 &  63.93 \\
\hline
\end{tabular}
%}
\label{tab:FinetuneExample}
\end{table}

Table~\ref{tab:DigitsFivefine-tune} and Table~\ref{tab:Office31fine-tune} show the comparison between with and without fine-tuning the source classifiers by the distilled source samples on Digits-five and Office31 datasets, respectively. It is clear that without distilling, the adaptation performance drops in most cases. For example, we can achieve 0.3\% and 0.5\% average accuracy improvements by source distilling on Digits-five and Office31 datasets. This confirms the validity of distilling the sources, since the selected source samples are more similar to the target ones and the fine-tuned classifier can enhance the transferability.

To better demonstrate the effectiveness of source distilling, we give an example of Wasserstein Distance based ADDA method before and after distilling on the Digits-five dataset when sy is set as the target domain and the others as source domains. As shown in Table~\ref{tab:FinetuneExample}, we find that the performance gains of source distilling vary across different sources. For the sources with larger domain discrepancies to the target, \textit{e.g.} mt to sy and up to sy, source distilling may yield higher improvement (2.5\% and 2.1\%, respectively), while the improvement is not that obvious for the sources with smaller discrepancy to the target, \textit{e.g.} sv to sy (0.1\%), mm to sy (0.2\%). This is reasonable because when one source domain is far away from the target, the distilled samples can lead the classifier closer to target domain. If the source is already very similar to the target, the influence of distilled samples will be not that obvious.

\begin{figure}[!t]
\begin{center}
\centering \includegraphics[width=0.95\linewidth, height= 7cm]{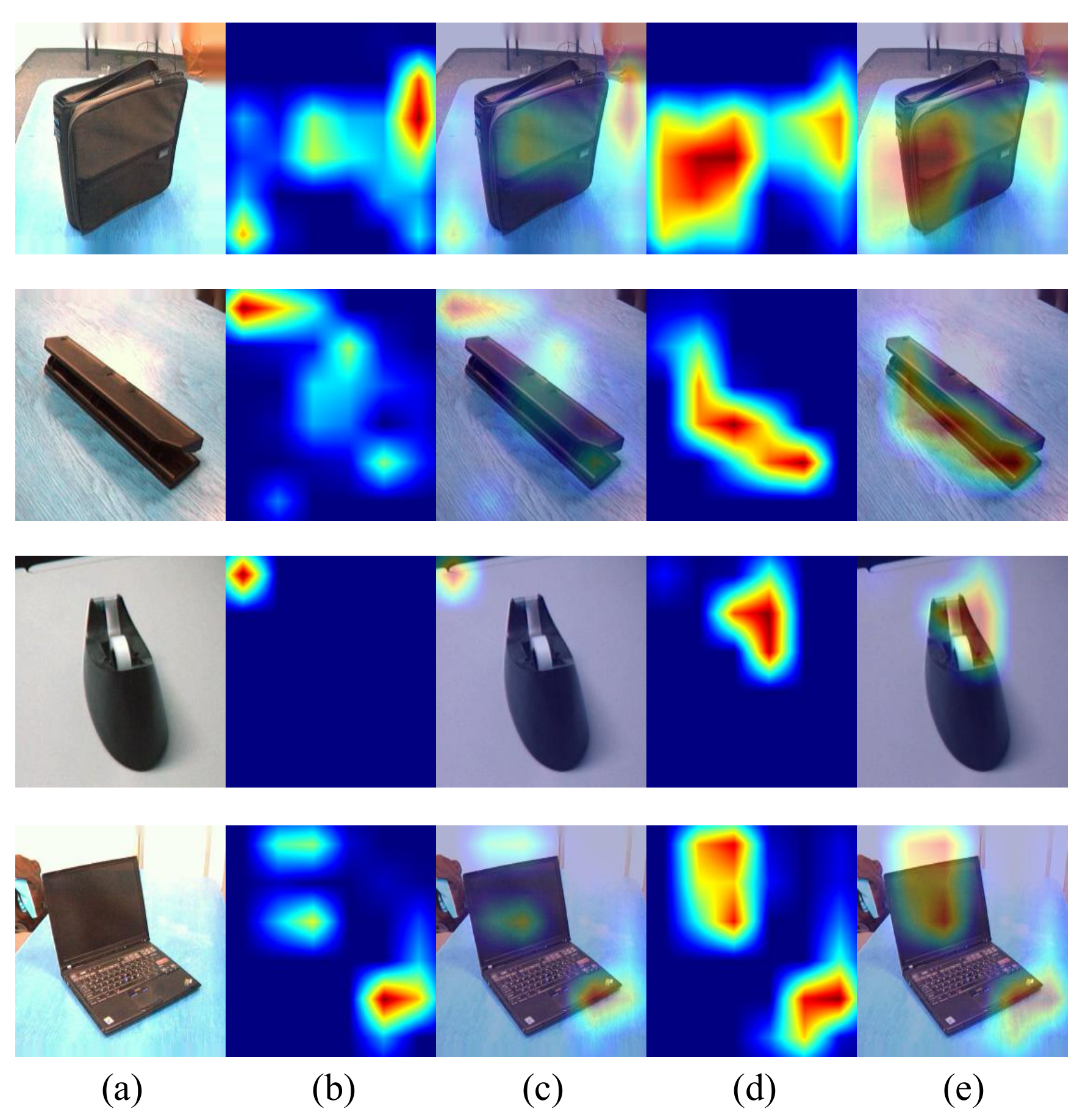}
\caption{Comparison of the attention maps before and after adversarial training on Office-31 dataset. From left to right: (a) original image; (b) attention map before adversarial training; (c) image with attention map before adversarial training; (d) attention map after adversarial training; (e) image with attention map after adversarial training. Brighter regions indicate more attention. Comparison shows the attention shifts to more discriminative regions of the image after adversarial training. Best viewed in color.}
\label{fig:heat1}
\end{center}
\end{figure}

\subsubsection{Model Interpretability} 

In order to show the interpretability of our model, we use the heat map generated by the Grad-Cam algorithm~\cite{gradcam2017iccv} to visualize the attention before and after our proposed domain adaptation method. As illustrated in Figure~\ref{fig:heat1}, we observe that after the domain adaptation: the attentions generated by our model can better focus on the more ``discriminative'' regions, which indicates that our model can pay more attention to the discriminative regions of the objects for classification even though the background or view point are changed. Such observation verifies that our model learns the features that are more invariant to different domains, while they are discriminative for the desired learning task (\textit{i.e.} image classification).%; (2) for those images with the attention on the discriminative regions before adaptation, our model makes such attention become stronger.
%(1) for those images with the wrong attention before adaptation, our model successfully changes the attention  to the more ``discriminative'' regions, which indicates that our model has learned through adaptation to pay more attention to the discriminative regions of the objects for classification even though the background or view point are changed.

For example, the ring binder in the first row shows that before adaptation, the model focuses on a region in the background, instead of the central target object. However, after our domain adaptation, the model can correctly focus on the ring binder and thus is more discriminative for the classification. Similar observations can be found in the second and third rows. In the last row, we find that attention is enhanced on the discriminative regions of the object (the laptop) after our domain adaptation.

%we observe the enhancement of attention through adaptation. 
%For example, the ring binder in the first line shows that before adaptation, the extracted feature focuses on right of the image, which is the background, instead of the central target object. However, after our domain adaptation, the model can correctly focus on the ring binder and thus is more discriminative for the classification. Similar observations can be found in the second and third line. In the last line, we observe the enhancement of attention through adaptation. 
%In the second line of the left column, the attentions on the central object has enhanced through adaptation, which means our adapted model focuses on more discriminative features and thus leading to a better prediction result.
%On the other hand, in Figure~\ref{fig:heat2}, attention on the discriminative regions has been strengthened after our proposed domain adaptation. For example, as we can see in the fourth line of the left column, more attention is paid to the notebooks than a single notebook after adaptation.
%In the left sixth line, the red attention region extends to the bottleneck. Also, in the last line of the left column, the attention covers the whole bubble after adaptation, rather than merely a small batch.

\section{Conclusion}
\label{sec:Conclusion}
In this paper, we have proposed an effective multi-source domain adaptation approach MDDA. The separately pre-trained feature extractor and classifier for each source domain can sufficiently explore the discriminability of labeled source data. The adversarial discriminative-adaptation and source distilling aim to match the target feature distribution to the source ones and to fine-tune the pre-trained classifiers. A novel weighting strategy is designed to jointly combine the predictions from different source classifiers. The extensive experiments conducted on Digits-five and Office-31 benchmarks demonstrate that MDDA achieves 3.3\% and 0.4\% performance improvements as compared to the state-of-the-art multi-source domain adaptation approaches (\textit{i.e.} DCTN) for digit and object classification. In future studies, we plan to extend the MDDA model to more challenging vision tasks, such as scene segmentation. We also aim to investigate methods that can combine generative and discriminative pipelines for multi-source domain adaptation.

\section{ Acknowledgments}
This work is supported by Berkeley DeepDrive and the National Natural Science Foundation of China (No. 61701273).

\bibliographystyle{aaai}
\scriptsize\bibliography{egbib}

\end{document}